\documentclass[a4paper]{article}

\usepackage{graphicx}
\usepackage{xcolor}

\usepackage[sort&compress,comma,authoryear]{natbib}

\usepackage[linesnumbered,ruled,vlined]{algorithm2e}

\SetCommentSty{mycommfont}

\SetKwInput{KwInput}{Input}                
\SetKwInput{KwOutput}{Output}              

\begin{document}

\title{Differentially Private Generation of Small Images}
\author{J. Schwabedal, P. Michel, and M. S. Riontino\\ Pionic, Berlin, Germany\\~\\
\footnotesize{e-mail: jschwabedal@gmail.com}\\
\footnotesize{code: github.com/jusjusjus/noise-in-dpsgd-2020/tree/v1.0.0}}
\date{\today}

\maketitle

\noindent\textbf{Abstract.}
We explore the training of generative adversarial networks with differential privacy to anonymize image data sets.
On MNIST, we numerically measure the privacy-utility trade-off using parameters from $\epsilon$-$\delta$ differential privacy and the inception score.  Our experiments uncover a saturated training regime where an increasing privacy budget adds little to the quality of generated images.
We also explain analytically why differentially private Adam optimization is independent of the gradient clipping parameter.  Furthermore, we highlight common errors in previous works on differentially private deep learning, which we uncovered in recent literature.  Throughout treatment of the subject, we hope to prevent erroneous estimates of anonymity in the future.

\section{Introduction}

Differential privacy is the \textit{de-facto} standard for anonymized release of statistical measurements \citep{Dwork2014}.  The method information-theoretically limits how much of any individual example in a data set is leaked into the released statistics.  Risks of privacy infringement remain therefore bounded independent of the infringement method.  This independence is key as increasingly powerful expert systems are trained on cross-referenced statistics, which may lead to accidental, inconceivably intricate, and hard-to-detect infringements.  In this work, we explore a method able to anonymize not only derived statistical measurements, but the underlying raw data set with the same differential-privacy guarantees.  We follow others in training generative adversarial neural networks (GANs) with differential privacy.

\citet{Abadi2016} proposed a method of training neural networks with differential privacy in their seminal paper.  The authors proposed to make the gradient computations of stochastic gradient descent (SGD) a randomized mechanism by clipping the L-2 norm of parameter gradients in each example, and by adding random Gaussian noise to the gradients.  
Using the same method, we show that it is possible to train generative adversarial networks (GANs) on high-dimensional images (for an overview see \citet{Gui2020}).  Their generator networks can then be used to synthesize image data that are of high utility for further processing, but have guaranteed privacy for the original data source.

%

%
Privacy-preserving training of GANs with DP-SGD has been attempted recently.  \citet{Beaulieu2017} trained a generator for labeled blood-pressure trajectories to synthesize anonymous samples from the SPRINT trial using the AC-GAN approach.
\citet{Zhang2018} devised the Wasserstein GAN with gradient penalty to train a generator for MNIST and CIFAR10.  To improve training with privacy, they grouped parameters according to their gradients and adjusted the clipping boundary for each group.
\citet{Xie2018} use the original Wasserstein GAN procedure wherein the critic's (discriminator's) parameters are clipped.  This in turn ensures that gradients are bounded, thus fulfilling the criteria to compute privacy bounds from the noise variance added to the gradients.

In the remaining, we discuss the prior works cited above and point out some important fallacies.  We then present or own analysis of differentially private synthetic data generated from the MNIST data set.
We discuss how DP parameters affect the quality of generated images, which we measure using the inception score.

\section{Methods}

\subsection{Differentially private stochastic gradient descent}
\label{sec:dp}

A randomized mechanism $h:D\to R$ satisfies \textit{($\epsilon,\delta$)-differential privacy} if the following inequality holds for any adjacent pair of data sets $d, d'\in D$ and $S\subset R$:
\begin{equation}
\label{eq:dp}
    P\left[h(d)\in S\right]\leq e^\epsilon P\left[h(d')\in S\right] + \delta~.
\end{equation}
The \textit{Gaussian randomized mechanism} adds to an $n$-dimensional mechanism $f$ random Gaussian noise with variance $C^2\sigma^2$, wherein $C$ is the L-2 sensitivity of the mechanism across neighboring data sets, and $\sigma$ is the noise multiplier.  Drawing vector-valued independent Gaussian variates $\xi\sim\mathcal{N}(0, C^2\sigma^2I_{n\times n})$, the mechanism is defined as
\begin{equation}
\label{eq:gaussianmechanism}
    h(x) = f(x) + \xi~.
\end{equation}

In the training of neural networks with stochastic gradient descent, the mechanism $f$ is the computation of parameter gradient update $g$, i.e.~the mean over parameter gradients $g_j$ from samples $j$ across a mini-batch $B$:
\begin{equation}
\label{eq:gradsum}
    g = \frac{1}{|B|}\sum_{j\in B} g_j
\end{equation}
Only one gradient $g_k$ differs in Eq.~(\ref{eq:gradsum}) between adjacent data sets.  Let us suppose we can limit the L-2 norm of individual gradients to $C$. Then $g$ has an L-2 sensitivity of $C/B$ across adjacent data sets.

Based upon this theory, \citet{Abadi2016} has formulated differentially private SGD, which we reproduce in Algo.~(\ref{algo:dpsgd}).  Note, that the gradient's L-2 norm of each example is clipped individually (line 5), and that independent and vector-valued Gaussian variates $\xi$ are added (line 6).  One may replace the simple descent step (line 7) with any higher order, or moment based algorithm of gradient descent such as RMSprop or Adam, for example \citep{RMSprop, Kingma2015}.
\begin{algorithm}
\label{algo:dpsgd}
\DontPrintSemicolon
  
    \KwInput{Examples $\{x_1,...,x_n\}$, neural-network parameters $\theta$, loss function $\mathcal{L}(\theta, x_i)$, learning rate $\eta$, noise scale $\sigma$, gradient norm bound $C$, random noise $\xi\sim\mathcal{N}(0,\sigma^2C^2I)$}
    \KwOutput{Differentially private parameters $\theta$}

    \For{$t<T$} {
        Sample a random minibatch $B$
        
        \For{$x_i\in B$} {
            Compute $g_i\leftarrow \nabla_{\theta_t}\mathcal{L}(\theta_t, x_i)$
            
            $g_i\leftarrow g_i/\max(1,||g_i||_2/C)$
        }
        
        $g\leftarrow\frac{1}{|B|}\left(\xi+\sum_i~g_i\right)$
        
        $\theta\leftarrow \theta - \eta_t g$
    }
\caption{Differentially private SGD \citep{Abadi2016}}
\end{algorithm}

\subsection{Differentially private generative adversarial training}

\citet{Zhang2018} observed that, when training a generative adversarial network for data release, the Gaussian mechanism can be confined either to the generator or to the critic.  They argue that applying the Gaussian mechanism only to the critic permits batch-right normalization techniques to be used in the generator.
In Algo.~(\ref{algo:dpgan}), we reproduce their differentially private Wasserstein GAN with gradient penalty.
Note the similarity between lines 9 and 10 in the critic step, and lines 5 and 6 in Algo.~(\ref{algo:dpsgd}).  These lines are responsible for implementing the Gaussian randomized mechanism.  Also, note that the generator training step does neither involve gradient clipping nor random perturbations.  It does also not depend on the data set as long as the schedule updating the learning rate $\eta_t$ is differentially private.  This includes early stopping.
\begin{algorithm}
\label{algo:dpgan}
\DontPrintSemicolon
  
    \KwInput{Examples $\{x_1,...,x_n\}$, neural-network parameters $\theta$ and $w$, learning rate $\eta$, noise scale $\sigma$, gradient norm bound $C$, random noise $\xi\sim N(0,\sigma^2C^2 I)$}
    \KwOutput{Differentially private parameters $\theta$}

    \For{$t<T$} {
        \For{$s=1,\dots,n_{\mathrm{critic}}$} {
            \tcc{differentially private critic step}
            Sample a random minibatch $B$
            
            \For{$x_i\in B$} {
                Sample $z$ from $P(z)$, and $\rho\in[0,1]$
                
                $y \leftarrow\rho x_i + (1-\rho) G(z)$
                
                $\mathcal{L}_i\leftarrow D(G(z))-D(x_i)+\lambda\left(||\nabla_xD|_y||_2-1\right)^2$ 
                
                $g_i\leftarrow \nabla_w\mathcal{L}_i$
                
                $g_i\leftarrow g_i/\max(1, ||g_i||_2/C)$
            }
            $g\leftarrow\frac{1}{|B|}\left(\xi+\sum_i~g_i\right)$
            
            $w\leftarrow w - \eta_t g$
        }

        \tcc{non-private generator step}
        Sample $|B|$ instances of $z_i$ from $P(z)$
        
        $g_t\leftarrow \frac{1}{|B|}\sum_{i=1}^{|B|}\nabla_\theta D(G(z_i))$
        
        $\theta\leftarrow \theta - \eta_t g_t$
    }
\caption{Differentially private WGAN-DP \citep{Zhang2018} }
\end{algorithm}

\subsection{Quantifying the loss of privacy}

Differentially private training of neural nets was formulated using the Gaussian mechanism, but its application is only useful if we are able to estimate tight upper bounds for the privacy lost.  This loss is quantified by the parameters $\epsilon$ and $\delta$ from Eq.~(\ref{eq:dp}).  Such an upper bound was derived by \citet{Mironov17} and \citet{Mironov19}, in which they use the theory of R\'enyi differential privacy (RDP).  Their analysis is technically complicated, and we will only sketch its elements here.

R\'enyi differential privacy (RDP) is formulated in terms of the R\'enyi divergence of two probability distributions $p$ and $q$:
\begin{equation}
    \label{eq:renyidivergence}
    D_\alpha(p~||~q)=\frac{1}{\alpha-1}\log \langle\left(\frac{p(x)}{q(x)}\right)^\alpha\rangle_{x\sim q}
\end{equation}
A mechanism $h:D\to R$ fulfills $(\alpha,\epsilon')$-RDP if, for all neighboring data sets $d,d'\in D$ and $S\subset R$, $h$ obeys the inequality:
\begin{equation}
    \label{eq:rdp}
    D_\alpha\left[P(h(d)\in S)~||~P(h(d')\in S)\right]\le\epsilon'
\end{equation}
\citet{Mironov17} also linked the two definitions of differential privacy (\ref{eq:dp}) and (\ref{eq:rdp}): Each mechanism satisfying $(\alpha,\epsilon')$-RDP also satisfies $(\epsilon,\delta)$-DP with
\begin{equation}
    \label{eq:dprelation}
    \epsilon=\epsilon'(\alpha) - \frac{\log\delta}{\alpha-1}~.
\end{equation}
One is free to choose $\delta$.  Then, one typically chooses the $\alpha$ that minimizes $\epsilon$.
\citet{Mironov19} give the details on how to compute $\epsilon'(\alpha)$ for one step of the sampled Gaussian randomized mechanism, stochastic gradient descent.  Across multiple steps, the epsilons add linearly.  Adding the values for $\epsilon'$ of each optimization step, we compute the privacy budget in terms of $\alpha$ and $\epsilon'$ and convert it to $\delta$ and $\epsilon$ using Eq.~(\ref{eq:dprelation}).
In sum, this gives us an upper bound for $(\epsilon,\delta)$-DP for our GAN training, that limits the privacy leaked from the data set into the generator.

\subsection{MNIST data set}

The MNIST dataset contains 70,000 labeled images of digits.  We used the 60,000 examples of its training data set to train GANs.  To optimize the classifier for the computation of inception scores, we also used the 10,000 examples in the test set as a validation set.

\subsection{Measuring the quality of generated images}

To assess the quality of generated images, we adopt the \textit{inception score} (IS): A classifier $K$ with classes $k$ generates a probability distribution $P(k|x)$ when applied to examples $x$ of a data set $X$.  The conditional probability is related to the marginal $P(k)=\langle P(k|x)\rangle_{x\sim X}$ using the Kullback-Leibler divergence:
\begin{equation}
    \label{eq:kldiv}
    \textrm{KL}(P~||~Q)=\langle\log{P} - \log{Q}\rangle_{x\sim X}
\end{equation}
The inception score $s(X)$ is defined as the exponential of the mean Kullback-Leibler divergence:
\begin{equation}
    \label{eq:is}
    s(X) = \exp\left[\sum_{k=1}^M\textrm{KL}(P(k|x)~||~P(k))/M\right]
\end{equation}

Note that the IS framework requires a classifier of high quality.   The score takes values between $1$ and the number of classes $M$.  It has been argued that the IS correlates well with subjective image quality because of a subjective bias towards class-distinguishing image features \citep{IS}.

\subsection{Architecture of the generative adversarial network}

We train Wasserstein generative adversarial networks using gradient penalty with Adam optimization.
The critic (discriminator) consists of three stridden convolutional layers with leaky ReLU activation functions of negative slope $0.2$.  The first layer has a number of filters, the \textit{capacity}, with a kernel size of $5$.  The number of filters doubles with each convolutional layer.
The generator starts with a $128$-dimensional Gaussian latent space that is processed by three convolutions that transpose the structure of the critic.  Padding is chosen to match the 28-by-28 pixel images of MNIST.
The network is trained in batches using Adam with $\beta_1=0.9$ and $\beta_2=0.5$ (default values in PyTorch and Tensorflow).

\section{Results}

\subsection{Review of prior work}

Differentially private stochastic gradient descent has been previously used to train generative adversarial networks \citep{Beaulieu2017, Zhang2018, Xie2018}.
\citet{Beaulieu2017} use the original GAN algorithm with a binary classification in the classifier.  They clip the gradients after averaging, but not the parameters (W-GAN step).  In Methods, they write:
\begin{quote}
  ... we limit the maximum distance of any of these [optimization] steps and then add a small amount of random noise.  
\end{quote}
As we outlined in Sec.~\ref{sec:dp}, limiting each step, i.e.~clipping the average gradient, is not sufficient to grant differential privacy to each example.  Each contribution to the gradient needs to be clipped individually.  At the time of writing, this error also conforms with their published code (github.com/greenelab/SPRINT\_gan).
Results presented by \citet{Beaulieu2017} may still be correct because the authors use a batch size of one throughout the paper.

In the public code repository implementing the experiments reported by \citet{Zhang2018}, we have encountered a stray factor $1/\sqrt{|B|}$ in the computation of the noise.  At a batch-size of 64, the noise is, therefore, a factor eight too small to grant the differential privacy constraints computed from the reported clipping and noise multiplier.  Furthermore, the clipping is performed in a non-standard way:  the authors group sets of parameters -- e.g.~all biases -- and clip their gradients' L-2 norm separately.  The difference to regular clipping is probably proportional to the number of groups.  The clipped gradients are, therefore, about a factor of about 10 larger than when clipped in the standard way.  These programming and conceptual errors leave the applied privacy analysis inapplicable.

\citet{Xie2018} use the original W-GAN algorithm in which parameters are clipped.  They go on to show that bounded network parameters, images, and classifications result in bounded parameter gradients thus fulfilling the DP conditions.  They compute the bound $c_g$, and use it for their DP algorithm.  Unfortunately, the authors neither write how they scale noise with $c_g$, nor do they publish their algorithm in the public repository (github.com/illidanlab/dpgan).

\subsection{Empirical exploration of differentially private synthetic data generated from MNIST}

To evaluate the inception score, we trained a classifier on the ten classes of the MNIST training data set.  On the test set, our trained classifier achieved an accuracy of 99.6\%, which is comparable to the state of the art for an individual neural-network classifier.
We also computed the inception score of the original data sets and achieved $s=9.73\pm 0.05$ (train set), and $9.61\pm 0.04$ (test set).

We used this classifier to compute the inception scores reported in the following sections.
Specifically, we describe our findings of how the gradient clipping, the noise multiplier, and the network capacity affected the privacy-utility relationship between privacy parameter $\epsilon$, and inception score IS.  In the computation of $\epsilon$, we set $\delta=10^{-5}$ to align our results with other publications that use the same value.  Note that there is still no consensus how to choose $\delta$, optimally.  In all figures, we counterposed the results with a non-anonymous GAN training.  We mark its maximum inception score, i.e.~$\textrm{IS}_{\textrm{max}}=8.51$, by a horizontal dashed line.  All experiments were done with the public code repository at "github.com/jusjusjus/noise-in-dpsgd-2020/tree/v1.0.0". 

\subsubsection{Adam is almost independent of gradient clipping}

For Adam optimization, "[...] the magnitudes of parameter updates are invariant to rescaling of the gradient" \citep{Kingma2015}.
Specifically, Adam tracks running means of the value $m_t$ and square $v_t$ of incoming averaged gradients $g_t$ at time step $t$.  The parameter update $\Delta p_t$ is normalized by these running means (some details omitted for clarity):
\begin{eqnarray}
    \label{eq:adam}
    m_t & = & \beta_1 m_{t-1} + (1-\beta_1) g_t \nonumber \\
    v_t & = & \beta_2 v_{t-1} + (1-\beta_2) g^2_t \\
    \Delta p_t & = & \eta \frac{m_t}{\sqrt{v_t}} \nonumber
\end{eqnarray}
Let us choose $C$ smaller than all individual gradient L-2 norms throughout the whole training process.  This is possible if we assume that the gradients are non-zero.  Then we can rewrite $g_t=C(\hat{\xi}+g_t/||g_t^2||)$, wherein $\hat{\xi} = \xi/C$ is independent of $C$.  After entering the expression for $g_t$ in Eqns.~(\ref{eq:adam}), $C$ cancels if we replace $m_0\to Cm_0$ and $v_0\to C^2v_0$.  For such small values of $C$, the training becomes $C$-independent due to the normalization property of Adam optimization.

\begin{figure}[ht]
    \centering
    \includegraphics[width=0.9\textwidth]{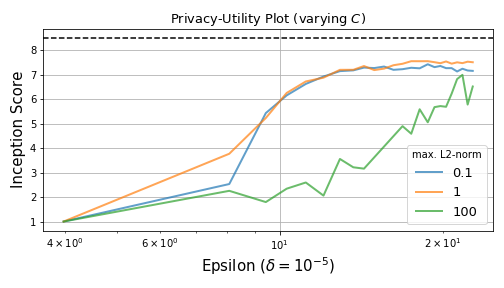}
    \caption{Privacy-utility plot for different gradient L2-norm clips $C$.  For $C=0.1$ and $C=1$ we observed comparable IS, whereas for $C=100$, IS as a function of $\epsilon$ was systematically reduced.}
    \label{fig:clip_dependency}
\end{figure}
Adam's rescaling property, thus, divides L-2 clipping constant $C$ into two regimes set by the smallest gradient norm in the data set.  (i) For smaller values of $C$, all gradients are clipped equally.  (ii) For larger values, the individual gradient norm weights the gradient sum over the mini-batch.

Empirically we found that regime~(i), in which $C$ is chosen arbitrarily small, showed the largest inception score.  Larger values of $C$ led to smaller values of the inception score.
We choose $C=1$ in the following, which was below the per-example gradient norm in our experiments.

\subsubsection{Critically large noise multipliers}

The noise multiplier $\sigma$ is inversely related to the signal-to-noise ratio in the gradients.  One may, therefore, expect that large $\sigma$ are detrimental to learning as optimal parameter values are escaped by random perturbations.

\begin{figure}[ht]
    \centering
    \includegraphics[width=0.9\textwidth]{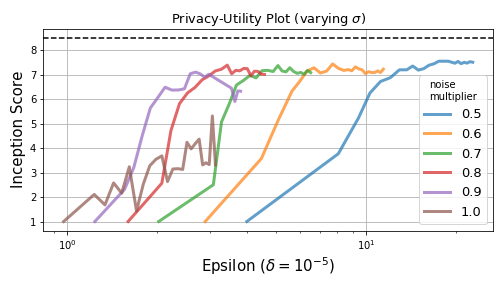}
    \caption{Privacy-utility plot for different noise multipliers $\sigma$.  At $\sigma<0.9$, we observed a plateau of IS that was mostly shifted to larger $\epsilon$ for smaller $\sigma$.  At $\sigma=1.0$, we observed another regimen in which training led to much lower values of IS with shallow gains during continued training.}
    \label{fig:sigma_dependency}
\end{figure}

We trained generators with DP-Adam at $C=1$, for a variety of values of $\sigma$ while monitoring their inception score and $\epsilon$ (Eq.~(\ref{eq:dp})).  Throughout training, the score increased initially then to approach a maximal level.  We found that the maximal inception score showed little dependence on values of $\sigma<0.9$ reaching about $s=7.2$. For $\sigma>0.9$, the score showed a steep break-off only reaching about $s=3$ at optimal levels of the privacy budget (cf.~Fig.~\ref{fig:sigma_dependency}).

Herein we observe the existence of a critical noise multiplier beyond which a privacy-utility trade-off will likely be sub-optimal, as we discuss further below.

\subsubsection{Optimal network capacity}

We trained generators with DP-SGD at a variety of network capacities while monitoring their inception score and privacy loss $\epsilon$ (Eq.~(\ref{eq:dp})).
The score approached a capacity-dependent maximum with increasing steps.  We found that the maximal inception score was biggest for an intermediate network capacity of $32$, while larger and small capacities reached lower levels.

\begin{figure}[ht]
    \centering
    \includegraphics[width=0.9\textwidth]{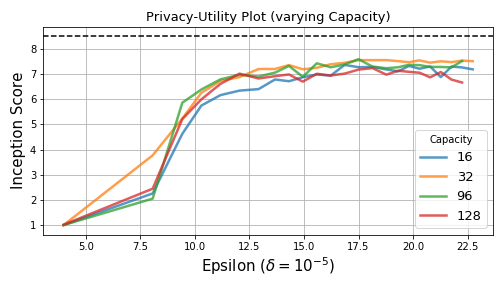}
    \caption{Privacy-utility plot for different network capacities.  We observed a consistent maximum for a network capacity of $32$ compared with the other tested capacities.}
    \label{fig:capacity_dependency}
\end{figure}

\section{Discussion}
Generative anonymization of data through differential privacy promises a quantifiable trade-off between the protection of individual privacy and the ability to use raw data sets for machine learning.
In this article, we explore the training of generative adversarial networks for image data under differential privacy constraints.
We found that some of the previous articles that explored this option showed technical and conceptual flaws \citep{Zhang2018, Beaulieu2017}.  In our introduction we provided a detailed workup of differential privacy in deep learning, which we hope will further clarify this intricate mathematical theory for future researchers.
It is in the interest of a practitioner to maximize the utility of the generator while staying within a specific privacy budget.   We explored this privacy-utility trade-off in the MNIST data set\footnote{Our work reproduces experiments initially published by \citet{Zhang2018}.  We do not compare these results to ours because of the aforementioned errors.}.
We uncovered two modes of DP-GAN training showing distinct characteristic  dependencies between privacy loss and image utility; an optimal one in which the utility steeply increased with spent privacy eventually reaching a plateau, and a sub-optimal one, wherein the slope was shallow (cf.~Fig.~\ref{fig:clip_dependency} at $C=100$ and Fig.~\ref{fig:sigma_dependency} at $\sigma=1$).  Sub-optimal privacy-utility characteristics crossed through optimal ones at low levels of the utility.  On the other hand, the plateau in optimal characteristics did not seem to dependent on the hyperparameters.  In Fig.~\ref{fig:sigma_dependency}, for example, we observed a stable plateau over an order of magnitude in $\epsilon$, and for different values of the noise multiplier.  We also found that the break-off between optimal and sub-optimal characteristics was abrupt, wherein increases in $\sigma$ or $C$ let to a sudden change in the observed characteristics.  We hypothesize that too small signal-to-noise ratios in anonymous gradient updates make the stochastic optimization process unable to uncover minima in the parameter space.  The hypothesis is consistent with break-offs upon increases in $\sigma$ and $C$.  It is also consistent with the large fluctuations in utility present in sub-optimal characteristics.
We also explained analytically why DP-GAN training with Adam becomes gradient-clipping independent for small values of the clipping constant.  This result generalizes to other methods of gradient descent with normalization.  Furthermore, we found a weak dependence of the utility plateau on the network capacity.  In the non-DP case, small capacities show reduced expressivity in the generator thus degrading the utility.  A reduction was not visible, however, at increased capacities (computations not shown).  When training with differential privacy, we found that increased capacity led to a systematic decrease in utility.  In these high-capacity networks, more terms enter the gradient L-2 norm thus enhancing the effect of the clipping.  We hypothesize that this is the main mechanism leading to a degraded utility in larger networks.

Our simple explorations are limited by the approximations with which we explore the privacy-utility relationship.  Privacy was indirectly measured as an upper privacy bound, and utility was indirectly measured with the inception score.  In future works, one should complement these metrics with direct measures of privacy through membership-inference attacks and classification scores on generated images, for example.  

In applications, privacy-utility plots could be a useful tool to tune parameters in a data anonymization workflow.  The method needs to be carefully adopted, however, because additional privacy loss incurs during hyperparameter optimization \citep{Abadi2016}, and the auxiliary classifier network we used needs to be trained with differential privacy as well.\\

In sum, we found that it is indeed possible to generate differentially private synthetic data set within a moderate privacy budget of $\epsilon\le10$.
However, we presume that the reduced utility for parameter-rich networks will be a major hurdle when training DP-GANs for larger, more nuanced image data sets than MNIST.  This could become a problem in particular, when close to a sub-optimal training regime in the signal-to-noise ratio of parameter gradients.

~\\
\noindent\textbf{Acknowledgements.}  We thank Dr.~Ilya Mironov for guiding our understanding of privacy-preserving deep learning.
This work was funded by the European Union and the German Ministry for Economic Affairs and Energy under the EXIST grant program.

\bibliographystyle{plainnat}
\bibliography{bib}
\end{document}